\documentclass{article}

\usepackage[nonatbib, preprint]{neurips_2023}

\usepackage[utf8]{inputenc} 
\usepackage[T1]{fontenc}    
\usepackage{hyperref}       
\usepackage{url}            
\usepackage{booktabs}       
\usepackage{amsfonts}       
\usepackage{nicefrac}       
\usepackage{microtype}      
\usepackage{xcolor}         
\usepackage{multirow}
\usepackage{array}
\usepackage{graphicx}
\newcolumntype{M}[1]{>{\centering\arraybackslash}m{#1}}

\title{Continuous Training and Fine-tuning for Domain-Specific Language Models in Medical Question Answering}

\author{%
  Zhen Guo\\
  MIT EECS\\
  Cambridge, MA 02139 \\
  \texttt{zguo0525@mit.edu} \\
  \And
  Yining Hua \\
  Harvard School of Public Health \\
  Boston, MA 02139 \\
  \texttt{yining\_hua@hms.harvard.edu} \\
}

\begin{document}

\maketitle

\begin{abstract}
Large language models exhibit promising general capabilities but often lack specialized knowledge for domain-specific tasks. Developing domain experts from a base model enables a range of applications without prohibitive training costs. This work demonstrates a method using continuous training and instruction fine-tuning to rapidly adapt Llama 2 base models to the Chinese medical domain. We first conduct continuous training on 1B tokens from Chinese medical references to teach relevant vocabulary and knowledge. The models are then fine-tuned on 54K examples sourced from the Chinese National Medical Licensing Examination. Experiments on Chinese medical data confirm the effectiveness of this approach, producing a model comparable to GPT-3.5-turbo while using way less computational resource. The resulting domain-specific model could be useful for various Chinese medical applications. More broadly, this provides a template for domain-specific training of large language models in areas where pre-trained models lack the required expertise, such as law, science, and engineering.
\end{abstract}

\section{Introduction}
Large Language Models (LLMs) have demonstrated impressive performance on a wide range of general language tasks in recent years. Models like GPT-3, PaLM, and Bard have achieved state-of-the-art results on applications including translation, question answering, and program synthesis~\cite{brown2020language, chowdhery2022palm, ouyang2022training, zhao2023survey}, enabled by massive training data, extensive pre-training, and rapid scaling~\cite{kaplan2020scaling, hoffmann2022training}. However, despite exhibiting excellent performance in the general domain, LLMs often fail to capture the specialized expertise required for fields such as medicine, law, or science~\cite{gu2021domain, hendrycks2021measuring}. As a result, LLMs are subject to poor performance on domain-specific applications, such as medical question answering or legal search~\cite{thirunavukarasu2023large, ahn2023impending, yue2023disclawllm, nguyen2023brief}.

Training domain-specific LLMs from scratch poses massive computational challenges, which are prohibitively expensive for most organizations~\cite{touvron2023llama}. In response, this work presents a pragmatic methodology for the rapid adaptation of general language models to specialized domains using continual training~\cite{gururangan-etal-2020-dont, ke2022continual, gupta2023continual} and instruction fine-tuning strategies. We focus on the healthcare and biomedicine domain, where the demand for proficient LLMs is hampered by stringent data privacy regulations and the limited computational resources typical of healthcare and biomedical labs~\cite{jbi_dl, PPR685311}. Concurrently, recognizing the linguistic inequity in NLP~\cite{bird2020decolonising, ijcai2023p0698} - primarily skewed towards English - we focus on Chinese, thus broadening the global utility and inclusivity of AI applications in healthcare. This dual challenge elevates the task difficulty and therefore presents a valuable evaluation for our proposed methodology. 

In detail, we develop a Chinese medical LLM based on the open-source Llama model~\cite{touvron2023llama}. The adaptation process is carried out in two stages. Initially, continual pretraining is employed on a corpus of 1 billion Chinese medical tokens derived from question-answer pairs. This aims to teach the model general medical knowledge. Subsequently, the model is fine-tuned on authentic medical licensing exam data with Alpaca format~\cite{alpaca} for medical question-answering.

The model developed with our method demonstrates capabilities comparable to those of GPT-3.5 but using significantly reduced computational resources for training. Compared to the original models, Llama-2-13B fine-tuned on CMExam with the question reasoning increased from 14.1\% to 38.8\% in F1 score, while Chinese-Llama-2-13B increased from 23.8\% to 44.1\%. Furthermore, additional gains were achieved with continuous training on 1 billion Chinese medical tokens before fine-tuning, with Llama-2-13B and Chinese-Llama-2-13B reaching F1 scores of 43.3\% and 45.7\% respectively. This result highlights the effectiveness of leveraging continual learning and fine-tuning to impart domain expertise and extended vocabulary into general models, thus presenting a viable framework for constructing specialized models for the biomedical domain. Moreover, the methodology presented in this work could be extended to other specialized domains. Examples include law, science, and engineering, where base models typically lack domain-specific knowledge.

\section{Method}

\subsection{Base models}
This work utilizes two base language models. The first is Llama 2 13B, which is pre-trained on 2 trillion tokens with a 4096 context length~\cite{touvron2023llama}. The second base model is a Chinese version of Llama 2 13B, initialized from the original Llama 2 and continuously trained on a mixture of Chinese and English corpora~\cite{li-etal-2022-csl, zhao2023tencentpretrain}. The Chinese data with sources such as encyclopedias and scientific papers provides knowledge about the Chinese language. The English sources, including datasets like SlimPajama~\cite{shen2023slimpajamadc} and RefinedWeb~\cite{penedo2023refinedweb}, maintain existing knowledge in the base Llama model.

\subsection{Continual training dataset}
The continual training dataset includes more than 364,000 question-answer pairs extracted from Chinese medical encyclopedias and expert articles online. These data are part of the larger Huatuo-26M dataset~\cite{li2023huatuo26m}, which contains around 1 billion tokens. The encyclopedia excerpt focuses on formal medical terminology and topics like andrology, gynecology, and infectious diseases rather than consultation records. The continuous training dataset could easily be expanded by adding more medical domain data, from open-sourced datasets in HuggingFace or proprietary datasets~\cite{luo2022biogpt, anil2023palm, singhal2023expertlevel}, given sufficient computational resources.

\subsection{Fine-tuning dataset}
CMExam is a large-scale Chinese medical exam dataset comprising more than 54,000 multiple-choice questions that span a wide range of medical knowledge. The questions cover 27 disease categories and 36 clinical departments, serving as a valuable resource for advancing medical language understanding~\cite{liu2023benchmarking}. CMExam also includes explanations for most of the questions, providing additional knowledge for fine-tuning LLMs by enabling them to incorporate reasoning capabilities during the training process.

 While this work focuses on evaluating medical knowledge, CMExam could potentially be combined with other resources, like dialogues from 
 Li et al.'s work\cite{li2023chatdoctor}, to fine-tune conversational agents for patient-doctor interactions. However, best practices for evaluating AI chatbots remain an open research area, as discussed in \cite{vicuna2023}.

\subsection{Evaluation datasets}

To evaluate the model's knowledge of the Chinese medical domain, we perform greedy decoding with a maximum generation length of 256 tokens on the 6,811 question-answer pairs in the CMExam test set ~\cite{liu2023benchmarking}. For base models not fine-tuned on the CMExam training set, including Llama-13B and Chinese-Llama-13B, we use a custom script to extract answers from the model outputs. The script first attempts to directly match the answer options in the model output. In case of failure, it performs a case-insensitive search for any answer strings provided in the answer options. As a last resort, the Levenshtein distance is used to find the closest matching answer to the first line of model prediction. For all other models, we use the same hard match evaluation as the original CMExam paper, comparing the ground truth answer to the predicted answer.

To assess catastrophic forgetting of general knowledge~\cite{Kirkpatrick_2017}, we use MMLU ~\cite{hendrycks2021measuring} for English and CMMLU ~\cite{li2023cmmlu} for Chinese, with 5-shot prompting for both. These general knowledge benchmarks allow us to measure whether medical fine-tuning impacts the retention of broader world knowledge in LLMs. C-Eval~\cite{huang2023ceval} could also provide valuable information, we did not include it in this study due to installation issues with our cluster.

\subsection{Hyperparameters for training}

We continuously train and fine-tune the models using the HuggingFace Transformers library on a cluster consisting of 20 nodes each with 6 NVIDIA V100 32GB GPUs, with Fully Sharded Data Parallelism (FSDP) to distribute the model across nodes~\cite{zhao2023pytorch}. The key hyperparameters are summarized in Tab.~\ref{hyperparameters-table}.

\begin{table}[h!]
  \caption{Hyperparameters for Training}
  \label{hyperparameters-table}
  \centering
  \begin{tabular}{ll}
    \toprule
    Hyperparameter Name & Value \\
    \midrule
    Learning rate & \(2 \times 10^{-5}\) \\
    Batch size & 120 \\
    max seq length & 4096\\
    Number of epochs & 1 \\
    Warmup ratio & 0.03 \\
    \bottomrule
  \end{tabular}
\end{table}

Techniques such as mixed precision fp16~\cite{micikevicius2018mixed}, gradient checkpointing~\cite{chen2016training}, and AdaFactor optimizer~\cite{shazeer2018adafactor} are used to improve training efficiency.

\section{Results}
\subsection{Training with reasoning improves model performance}

We first conduct full-parameter fine-tuning on the two based models: Llama-2-13B and Chinese-Llama-2-13B. The fine-tuning data come from the CMExam training set, using answer-only examples or question-answer pairs with reasoning explanations. For models in the general domain, the performances of medical question-answering are suboptimal except for GPT-3.5-turbo and GPT-4. As shown in Table \ref{model-performance-table1}, training with explanations in the answer improves the test accuracy for both models, even when evaluating accuracy alone, suggesting the values of reasoning data in fine-tuning. 

When fine-tuned solely on the CMExam training data, Chinese-Llama-2-13B achieves comparable performance to GPT-3.5-turbo. In the CMExam paper, a fine-tuned Llama-7B model obtained a lower accuracy of just 18.3\%. This reduced performance could be attributed both to the smaller model size and to the use of prompt tuning~\cite{li2021prefix} and low-rank adaptation~\cite{hu2021lora}. While these approaches are parameter-efficient, full-parameter fine-tuning generally enables better performance.

\begin{table}[h!]
  \caption{Performance Metrics for Fine-tuning}
  \label{model-performance-table1}
  \centering
  \begin{tabular}{ccccc}
    \toprule
    \multirow{2}{*}{Model Type} & \multirow{2}{*}{Model} & \multirow{2}{*}{More Finetuning Dataset} & \multicolumn{2}{c}{CMExam} \\
    \cmidrule(r){4-5}
    & & & Acc (\%) & F1 (\%) \\
    \midrule
    \multirow{5}{*}{General Domain} & GPT-3.5-turbo & - & 46.4 & 46.1 \\
    & GPT-4 & - & \textbf{61.6} & \textbf{61.7} \\
    & ChatGLM-7B & - & 26.3 & 25.7 \\
    & Llama-2-13B & - & 21.0 & 14.1\\
    & Chinese-Llama-2-13B & - & 26.7 & 23.8 \\
    \midrule
    \multirow{4}{*}{Medical Domain} & \multirow{2}{*}{Llama-2-13B} & CMExam\_answer\_only & 37.3 & 37.0 \\
    &  & CMExam\_with\_reasoning & 39.3 & 38.8 \\
    & \multirow{2}{*}{Chinese-Llama-2-13B} & CMExam\_answer\_only & 42.9 & 42.7 \\
    &  & CMExam\_with\_reasoning & \textbf{44.5} & \textbf{44.1} \\
    \bottomrule
  \end{tabular}
\end{table}

\subsection{Continual training improves domain knowledge at a cost}
\begin{figure}[h!]
  \centering
  \includegraphics[width=0.55\textwidth]{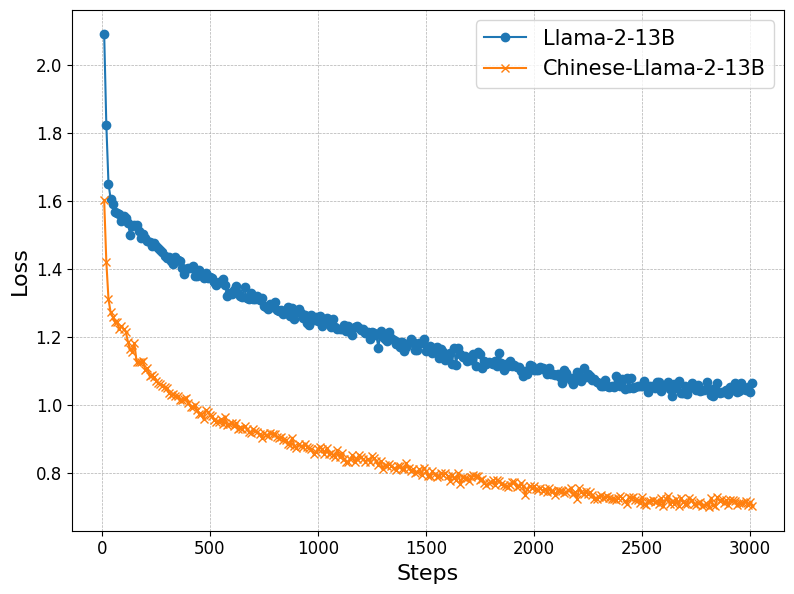}
  \caption{Continual training loss on encyclopedia QA dataset}
  \label{fig:train_loss}
\end{figure}

Next, we show the training loss in Fig.~\ref{fig:train_loss}. The perplexity loss decreases gracefully despite the distribution shift from pre-training to continual training data. Notably, our continual training is around 1 billion tokens, much smaller than the 2 trillion tokens used for Llama-2 pre-training.

In Table~\ref{model-performance-table2}, we show the CMExam results on different continual training checkpoints. At each checkpoint, we also fine-tune the model with question explanations (reasoning) associated with the questions. A trend emerges where the model accuracy on CMExam tends to increase as continual training progresses, signifying th learning of medical domain knowledge. However, decreases are observed in the MMLU and CMMLU metrics over time, pointing to a phenomenon of catastrophic forgetting of general knowledge.  This is unsurprising since we did not mix in samples from the original pre-training data due to computational constraints. As a result, the model may have progressively replaced general knowledge with new knowledge in the medical domain. Overall, after 1 epoch of continuous training, the model exhibits only minor decreases in general knowledge, along with a small improvement in healthcare and biomedical understanding. This trend of forgetting general knowledge and gains in domain-specific knowledge could continue if training is extended over more epochs.

\begin{table}[h!]
  \caption{Performance Metrics for Fine-tuning Continuously Trained Checkpoints}
  \label{model-performance-table2}
  \centering
  \begin{tabular}{cccccc}
    \toprule
    \multirow{2}{*}{Model} & \multirow{2}{*}{Trained Checkpoint} & \multicolumn{2}{c}{CMExam} & \multicolumn{1}{c}{MMLU} & \multicolumn{1}{c}{CMMLU}\\
    \cmidrule(r){3-4} \cmidrule(r){5-5} \cmidrule(r){6-6}
    & & Acc (\%) & F1 (\%) & Acc (\%) & Acc (\%)\\
    \midrule
    \multirow{5}{*}{Llama-2-13B} & 0 & 39.3 & 38.8 & \textbf{58.6} & \textbf{42.7}\\
    & 750 & 43.4 & 43.0 & 57.2 & 42.1\\
    & 1500 & 43.1 & 42.8 & 57.2 & 42.1\\
    & 2250 & 43.6 & 42.9 & 56.4 & 42.0\\
    & 3000 & \textbf{43.8} & \textbf{43.3} & 57.0 & 41.8\\
    \midrule
    \multirow{5}{*}{Chinese-Llama-2-13B} & 0 & 44.5 & 44.1 & \textbf{54.4} & \textbf{46.0}\\
    & 750 & 44.7 & 44.7 & 54.4 & 45.7\\
    & 1500 & 45.5 & 45.1 & 54.0 & 45.1\\
    & 2250 & 45.9 & 45.5 & 53.1 & 45.3\\
    & 3000 & \textbf{46.0} & \textbf{45.7} & 53.1 & 45.2\\
    \bottomrule
  \end{tabular}
\end{table}

\subsection{Challenges in cross-lingual fine-tuning}

\begin{table}[h!]
  \caption{Performance Metrics for Fine-tuning with Mixed Datasets}
  \label{model-performance-table3}
    \centering
  \begin{tabular}{ccccc}
    \toprule
    \multirow{2}{*}{Model} & \multirow{2}{*}{Checkpoint} & \multirow{2}{*}{Finetuning Dataset} & \multicolumn{2}{c}{CMExam}\\
    \cmidrule(r){4-5}
    & & & Acc (\%) & F1 (\%)\\
    \midrule
    \multirow{2}{*}{Llama-2-13B} & 3000 & CMExam & 43.8 & 43.3\\
    & 3000 & CMExam+MedQA+MedMCQA & 42.8 & 43.4\\
    \midrule
    \multirow{2}{*}{Chinese-Llama-2-13B} & 3000 & CMExam & 46.0 & 45.7\\
    & 3000 & CMExam+MedQA+MedMCQA & 41.3 & 43.2\\
    \bottomrule
  \end{tabular}
\end{table}

Inspired by Med-PaLM 2~\cite{singhal2023expertlevel}, we attempt to further improve Chinese medical QA performance by mixing English and Chinese medical question-answering data for fine-tuning. Specifically, we combine the MedQA~\cite{jin2020disease}, MedMCQA~\cite{pmlr-v174-pal22a}, and CMExam training sets and use this mixture to fine-tune the continuously trained Llama-2 and Chinese-Llama-2 checkpoints.

However, contrary to expectations, this multilingual fine-tuning results in degraded performance on the CMExam benchmark, especially for the Chinese-Llama variants, as shown in Table~\ref{model-performance-table3}. This observation implies that although techniques such as ensemble refinement and instruction tuning provide gains for English medical QA with Med-PaLM 2, they do not transfer directly when applied to Chinese medical data.

Thus, it highlights the challenges of cross-lingual transfer learning. Although methods from English tasks can inform model development in other languages, adaptation is still needed to account for linguistic differences. The optimal training strategies likely vary across languages and use cases. Simply mixing diverse multilingual data does not guarantee improved performance. More research is required to determine robust techniques for cross-lingual transfer that can truly augment in-language training.

\section{Conclusion}
Our experiments show the promise of continual pre-training and fine-tuning for developing domain-specific language models, rather than training from scratch. However, catastrophic forgetting of general knowledge remains a challenge. Mitigating this trade-off through techniques like experience replay could enable efficient adaptation of models to new domains while retaining diverse capabilities. Further research into continual learning is needed to fully realize its advantages.

\section{Ethical statement}
Continual training introduces the risk of amplifying biases in newly incorporated datasets, necessitating periodic bias audits and fairness-aware algorithms. Fine-tuning specialized medical tasks requires rigorous validation by medical experts to mitigate the risk of disseminating inaccurate or misleading medical information. The model is designed to complement, not replace, human expertise in healthcare settings, and given its focus on the Chinese medical language, it is fine-tuned to be sensitive to linguistic and cultural nuances. Transparency and accountability are emphasized to ensure ethical deployment, and ongoing ethical oversight is maintained to align the project with principles of beneficence and nonmaleficence.


\bibliographystyle{unsrt}
\bibliography{reference}

\end{document}